\documentclass{article}
\usepackage{amsmath}
\usepackage{amsthm}
\usepackage{amssymb}
\usepackage{graphicx}
\usepackage{multirow}
\usepackage{morefloats}
\usepackage{color, palatino}
\usepackage{verbatim}
\usepackage{subfigure}
\usepackage{pdfpages}

\newtheorem{proof*}{Proof}

\begin{document}
\title{Computational role of eccentricity dependent cortical magnification}

\includepdf[pages={1}]{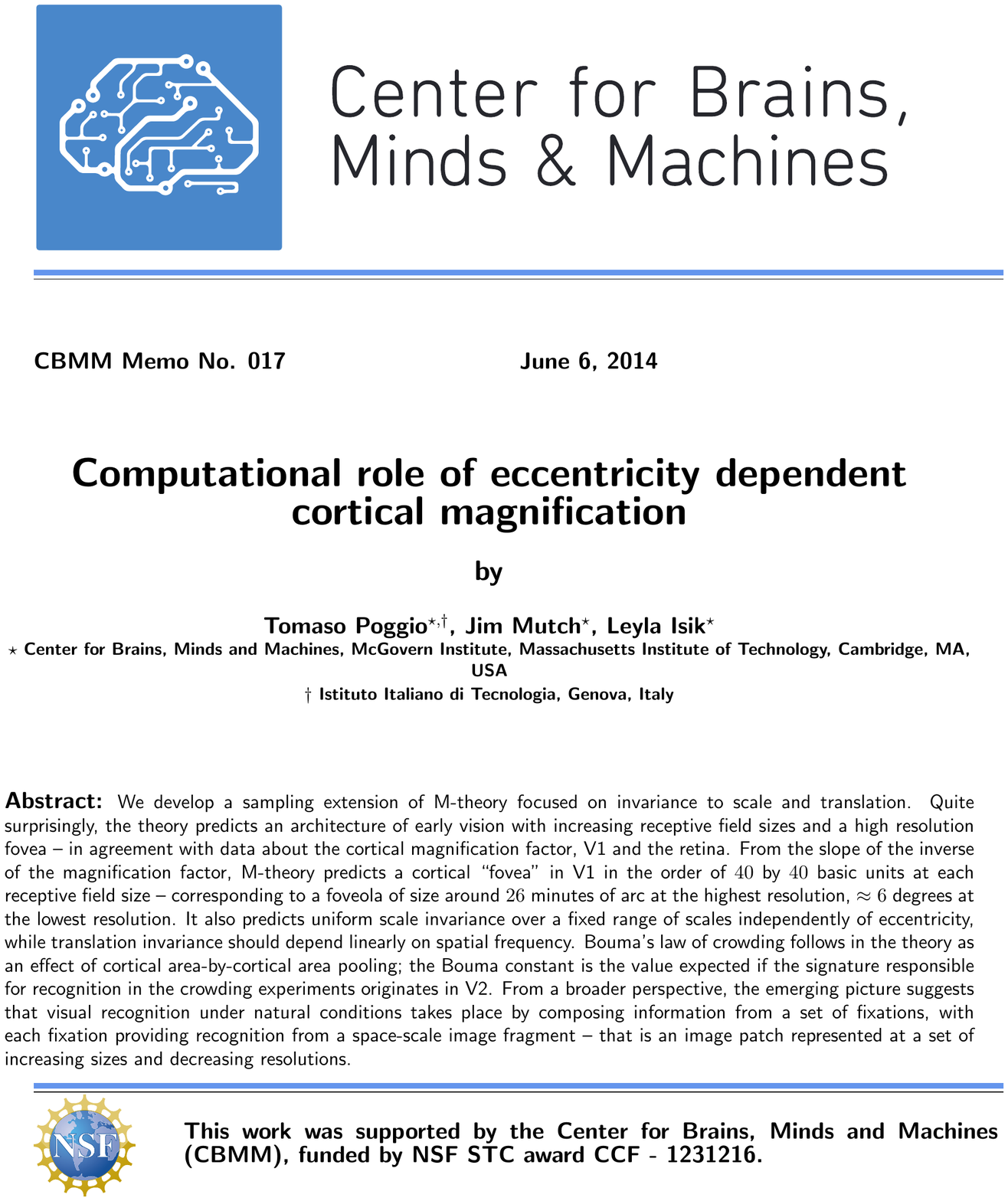}
\setcounter{page}{1}
% \toappear{...}

\maketitle

\begin{center}

\author{\normalsize{Tomaso Poggio$^{\star,\dagger}$, Jim
    Mutch$^{\star}$, Leyla Isik$^{\star}$}
\\
%\small \em $\star$  CBCL, McGovern Institute, Massachusetts Institute of Technology, Cambridge, MA, USA\\
\small \em $\star$  Center for Brains, Minds and Machines, McGovern Institute, Massachusetts Institute of Technology, Cambridge, MA, USA\\
\small \em $\dagger$ Istituto Italiano di Tecnologia,   Genova, Italy\\
%{\small \tt  \{tp,lrosasco\}@mit.edu}
}

%\author{Tomaso Poggio + Jim Mutch + Leyla Isik}
\end{center}
\vspace{0.5in}
{\bf Summary} {\it

\noindent We develop a sampling extension of M-theory focused on invariance to
scale and translation. Quite surprisingly, the theory predicts an
architecture of early vision with increasing receptive field sizes and
a high resolution fovea -- in agreement with data about the cortical
magnification factor, V1 and the retina.  From the slope of the
inverse of the magnification factor, M-theory predicts a cortical
``fovea'' in V1 in the order of $40$ by $40$ basic units at each receptive
field size -- corresponding to a foveola of size around $26$ minutes
of arc at the highest resolution, $\approx 6$ degrees at the lowest
resolution. It also predicts uniform scale invariance over a fixed
range of scales independently of eccentricity, while translation
invariance should depend linearly on spatial frequency. Bouma's law of
crowding follows in the theory as an effect of cortical
area-by-cortical area pooling; the Bouma constant is the value
expected if the signature responsible for recognition in the crowding
experiments originates in V2.  From a broader perspective, the
emerging picture suggests that visual recognition under natural
conditions takes place by composing information from a set of
fixations, with each fixation providing recognition from a space-scale
image fragment -- that is an image patch represented at a set of
increasing sizes and decreasing resolutions.

}

\vspace{0.25in}
\noindent
\large{\bf Motivation and plan}

\noindent A recent theory -- dubbed ``M-theory''\cite{MM2013} --
conjectures that the main computational task of the ventral stream of
visual cortex is to compute a representation of any new image in terms
of a signature which is invariant to transformations that have been
experienced in the past with other images and objects.  The conjecture
is motivated by theorems showing that invariant representations can
significantly reduce the sample complexity of a supervised learning
classifier. The magic theory proposes a family of algorithms for
unsupervised learning of an invariant representation.  The computation
of an invariant representation starts in the early visual areas with
affine transformations, mainly 2D translations and scale changes (and
some rotation). Here we show that an analysis of the sampling aspects
of the theory offers an intriguing interpretation of the role of
eccentricity-dependent size of receptive fields in V1, V2, V4 (and IT)
-- and of the linear dependency with eccentricity of the spatial
sampling of the photoreceptors in the primate retina. It also yields
several predictions.

This memo is divided into two parts: Part I analyzes eccentricity
dependence in the first stage of vision (eg V1, eg the first layer in
a hierarchical network), while Part II discusses plausible scenarios
for the higher visual areas (eg higher layers in the hierarchical
network). The main new results of Part I are:
\begin{itemize}
\item simultaneous invariance to translation and scale leads to an
  inverted, truncated pyramid as a model of the set of receptive
  fields in V1; this is a new alternative, as far as we know, to the
  usual smooth empirical fit of the cortical magnification factor data.

\item the model above predicts, as a special case, results from Anstis
  about recognition of letters at different eccentricities
  \cite{Anstis1974}.

\item the model also predicts that shift invariance depends on
  resolution wehereas scale invariance is uniform across eccentricities.

\item the size of the flat, high acuity foveal region, which we
  identify with the {\it foveola} can be inferred from the slope of
  the eccentricity dependent acuity. Existing neural data from macaque
  V1 suggest a foveola with a dimater of around $20'$ of arc.

\item scale invariance turns out to be more natural and complete than
  shift invariance. This is to be expected in organisms in which eye
  fixations can easily take care of shift transformations whereas more
  expensive motions of the whole body would be required to change the
  scale.

\end{itemize}

A knowledge of the main results of the magic theory is assumed. Much
of the results and discussion below is in computational terms without
reference to neuroscience. Biological implications and predictions are
presented in the final section.

\section{Part I: V1 simple cell receptive fields as stored transformations of
  templates}

The theory shows how the first step of the invariance computation may
consist of unsupervised learning of the set of transformations. This
is done by

\begin{enumerate}
\item storing template $t_k$ (which is an image patch and
could be chosen at random),
\item storing all its observed transformations (bound
  together by continuity in time), and
\item repeating the above process for a set of $K$ templates.
\end{enumerate}

The simplest and most common image transformations are similitude
transformations, that is shifts in position and uniform changes in scale.
Though the templates can be arbitrary image patches, we will assume
that V1 templates are Gabor-like functions (windowed Fourier transforms).

\begin{figure}\centering
\includegraphics[width = 0.8\textwidth]{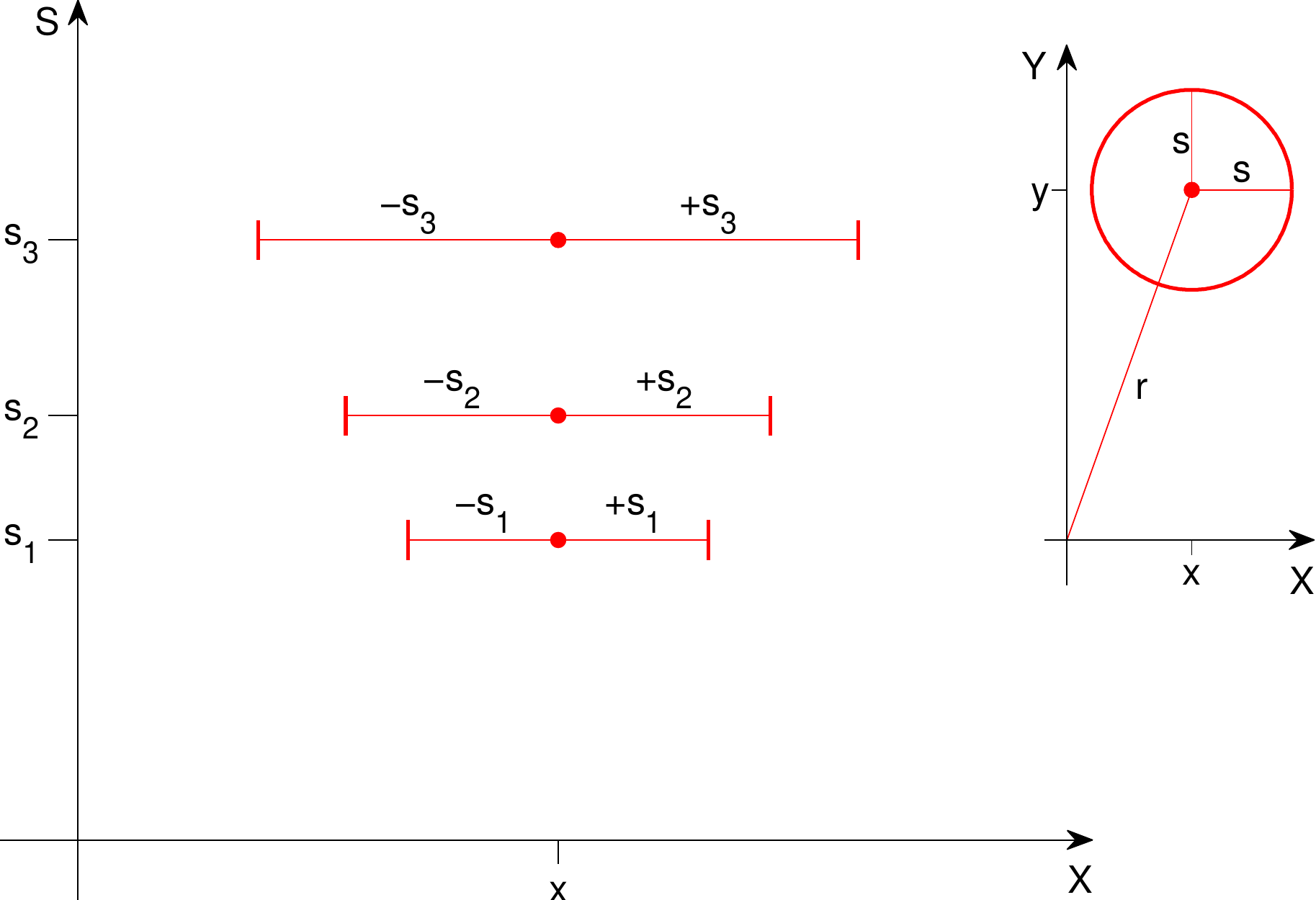}
\caption{{\it The $s,x$ plane, showing example templates
  of radius $s_1$, $s_2$, and $s_3$, all centered at position $x$.
  Note that while $s$ and $x$ are both measured in degrees of visual
  angle, in this plot the two axes are not shown to the same scale.
  Here, as in the rest of the paper, we show only one spatial
  coordinate ($x$); everything we say can be directly extended to the
  $x,y$ plane.  {\em Inset:} an example template of radius $s$ shown
  in the original $x, y$ plane.  $x$ in the $s,x$ plane corresponds to
  $r$ in the $x,y$ plane. We will assume later that the smallest
  template is the smallest simple cell in the fovea with a radius of
  around 40'' \cite{{Marretal1980}}.}}
\label{definitions}
\end{figure}

%{\bf Remark}

%Gabor functions have the remarkable propert that they are optimal in
%the sense that they maximize simultaneous invariance to scale and position.

\subsection{Geometry}

\begin{figure}\centering
\includegraphics[width = 0.5\textwidth]{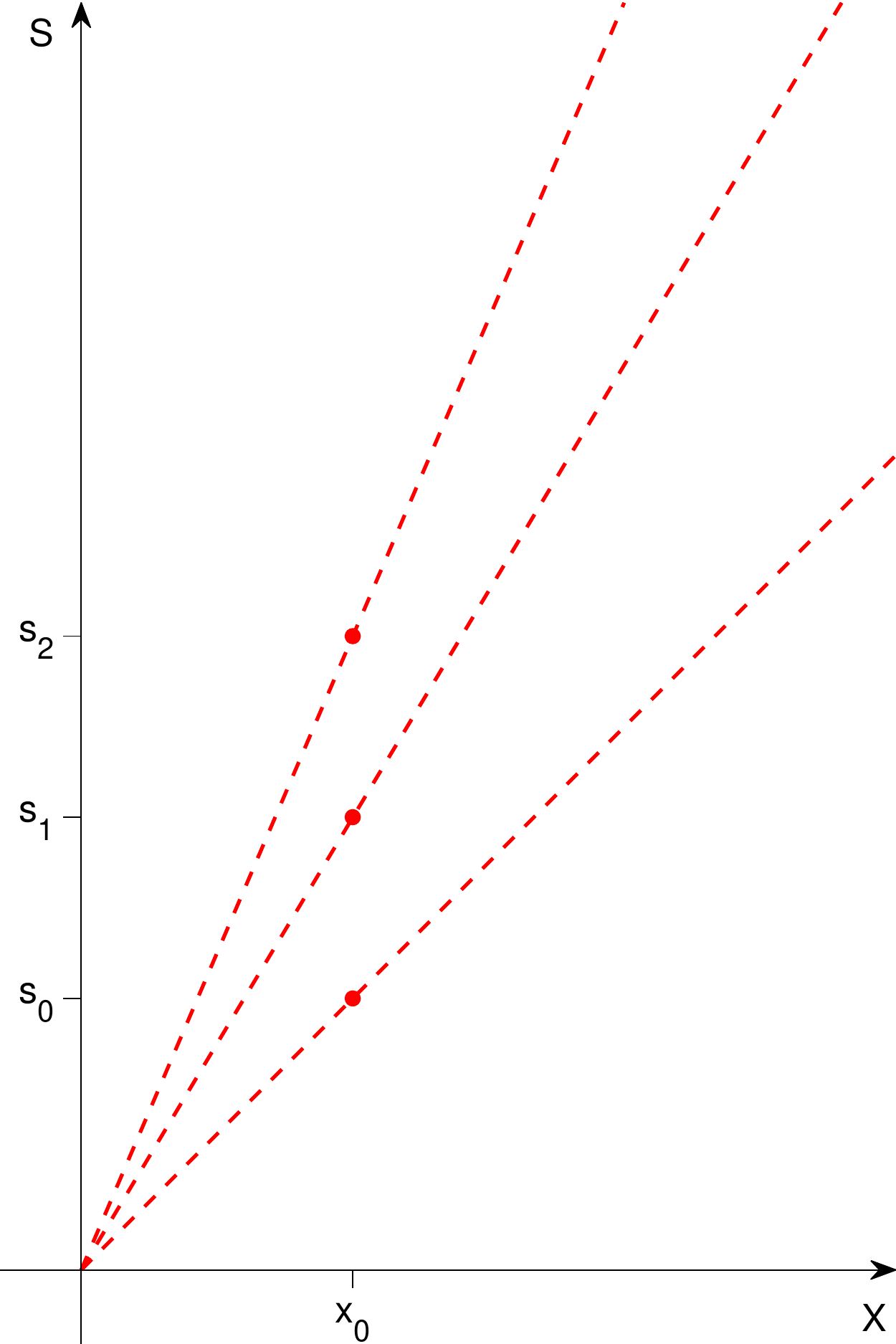}
\caption{\it For any fixed position $x_0$, the size of a pattern or
  template at that position determines the slope of its $s, x$
  trajectory under scaling centered at $x=0$. Note that the $s,x$
  trajectories under scaling are straight lines through the origin.}
\label{geometry}
\end{figure}

The geometry of scaling is of course independent of eccentricity
dependence in the retina or V1. Under scaling, a pattern exactly
centered at $0$ on the $x$ axis (eg the center of the fovea) will
increase in size without any translation while its boundaries will shift
in $x$; a pattern at say $x_0$ will increase in size and its center
will translate in the $s, x$ plane, according to $g_a I(x-x_0) =
I(\frac{x-x_0}{a})$, see Figures \ref{definitions} and \ref{geometry}.
In the $s,x$ plane the slope of the trajectory of a pattern under
scaling is a {\it straight line through the origin} with a slope that
depends on the size of the pattern $s$ and the associated position.

% JM: when do we say why we care mainly about *centered* scaling?
% (Because you can move your eyes.)

\subsection{Magic window instead of M scaling functions}

\begin{figure}\centering
\includegraphics[width = 0.7\textwidth]{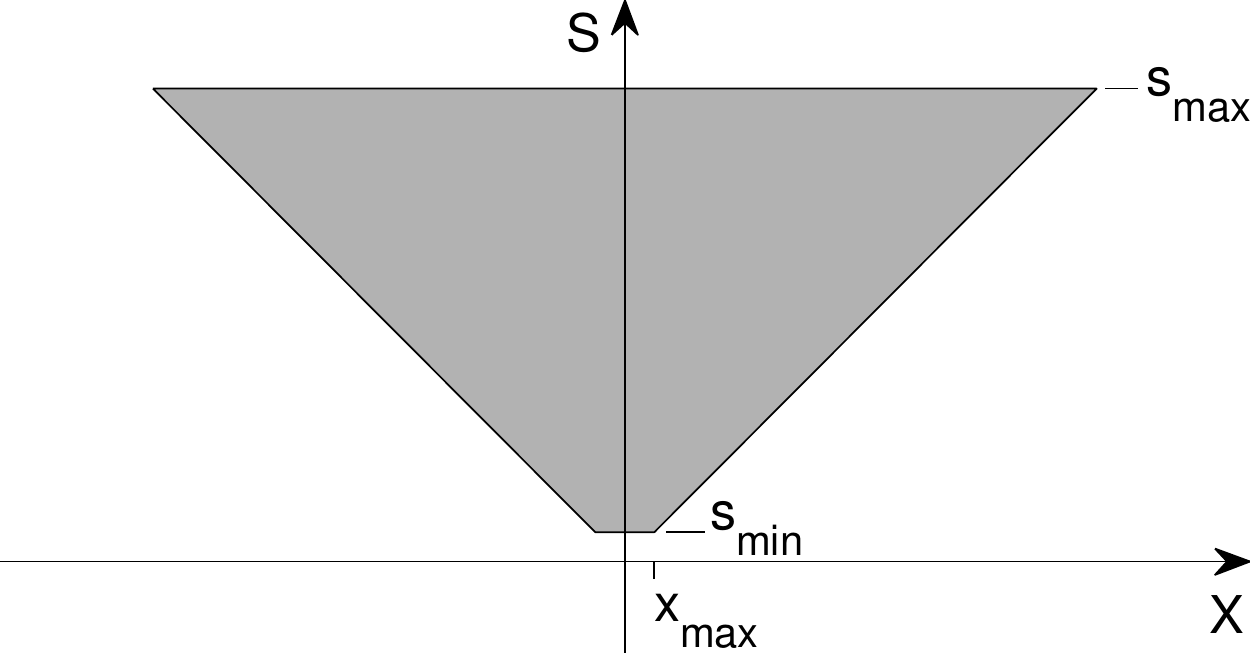}
\caption{\it Let us start with a template at $s_{min}$ and $x=0$. Its
  transformations under (bounded) shift at $s=s_{min}$ fill the line between $-x_{max}$
  and $x_{max}$. Its transformations under (bounded) scaling
  fill the space in the truncated, inverted pyramid shown in
  the figure. This is the space of bounded joint transformations in
  scale and space. The axes $s,x$ are here in the same units.}
\label{pyramid}
\end{figure}

Let us start with a template at $s=s_0, x=0$. $s_0$ could be for
instance the minimum possible receptive field size (given constraints
such as the optics). Transforming it by shifts in $x$ within $-x_0,
+x_0$ (the translation group, as the scale group, is only locally
compact) generates a set $S_0$. Suppose that we want to ensure that
what is recognizable at the highest resolution ($s_0$) remains
recognizable at all scales -- that is distances from the object -- up
to $s_{max}$. The associated scale transformations of the set $S_0$
yield the inverted truncated pyramid shown in Figure
\ref{pyramid}. Pooling over that set of transformed templates
according to the magic algorithm will give uniform invariance to all
scale transformations of a pattern over the range $(s_0, s_{max})$;
invariance to shifts will be at least within $(-x_0, x_0)$, depending
on scale. Note that the above process of observing and storing a set
of transformations of templates in order to be able to compute
invariant representations may take place at the level of evolution or
at the level of development of an individual system or as a
combination of both.

%The boundaries of the inverted
%pyramid are a window that corresponds to a minimum shift invariance
%between $x_{min}$ and $x_{max}$ and uniform scale invariance between
%$s_{min}$ and $s_{max}$.
The following definition holds: {\it the inverted truncated pyramid of Figure
\ref{pyramid} is the locus $S$ of the points such that their scaling between
$s_{min}$ and $s_{max}$ gives points in $S$; further all points $P$  between
$-x_0$ and $x_0$ are in $S$, eg $P \in S$ if

$$
g_s P \in S, s\in [s_{min}, s_{max}], S_0  \in S
$$

\noindent where $S_0$ consists of all points at $s_{min}$ between
$-x_0$ and $x_0$.
}

%The same truncated pyramidal region follows from the following process
%{\it 1) do shift transformations of a template at scale $s$ between
%  $-x_0$ and $x_0$ and 2) do scale transformations of all the points
%  so generated between $s_{min}$ and $s_{max}$.}

The region above follows naturally if scale invariance has a higher
priority than shift invariance. Alternatively, the complementary
constraint of uniform translation invariance for all points generates
in $s,x$ space a rectangular region between $x_{min}$ and $x_{max}$
and $s_{min}$ and $s_{max}$. In this case, points in the region do not
allow uniform scale invariance (in fact the scaling ranges from $0$ to
$s_{max} -s_{min}$) . The region of Figure \ref{pyramid} may be more
``natural'': we conjecture that the region of Figure \ref{pyramid} may
be optimal in terms of containing simultaneously maximum information
about scale and space (the open problem is to make the notion of
optimality precise). Notice that the best template in terms of
simultaneously maximizing scale and shift invariance is a Gabor
function (see \cite{MM2013}). Under scale and translation the Gabor
template originates a set of Gabor wavelets (a tight frame). In the
region of Figure \ref{pyramid} scaling (between $s_{min}$ and
$s_{max}$ ) each wavelet generates other elements of the frame.

Notice that the shape of the lower boundary of the inverted pyramid,
though similar to standard empirical functions that have been used to
describe M scaling, is different for small eccentricities. An example
of an empirical function (Cowey and Rolls, 1974 \cite{Cowey1974}, described in
\cite{Strasburger2011}) is $M^{-1} = M^0 (1+ax)$, where $M$ is the
cortical magnification factor, $M^0$ and $a$ are constants and $x$ is
eccentricity.

\subsection{Inverted pyramid scale-space image fragments}

A normal, non-filtered image may activate all or part of the Gabor
filters within the inverted truncated pyramid of Figure \ref{pyramid}.
The pattern of activities is related to a multi-resolution wavelet
decomposition of an image. We call this transform of the image
an``inverted pyramid scale-space fragment'' (``IP fragment'' in short)
and consider it as supported on a domain in the 2D space $x,s$ that is
contained in the inverted truncated pyramid of the figure. The
fragment corresponding to a bandpass filtered image should be a more
or less narrow horizontal slice in the $s,x$ plane. The term fragment
is borrowed from Ullman.

\subsection{Sampling the templates: specific assumptions}

\begin{figure}\centering
\includegraphics[width = 0.7\textwidth]{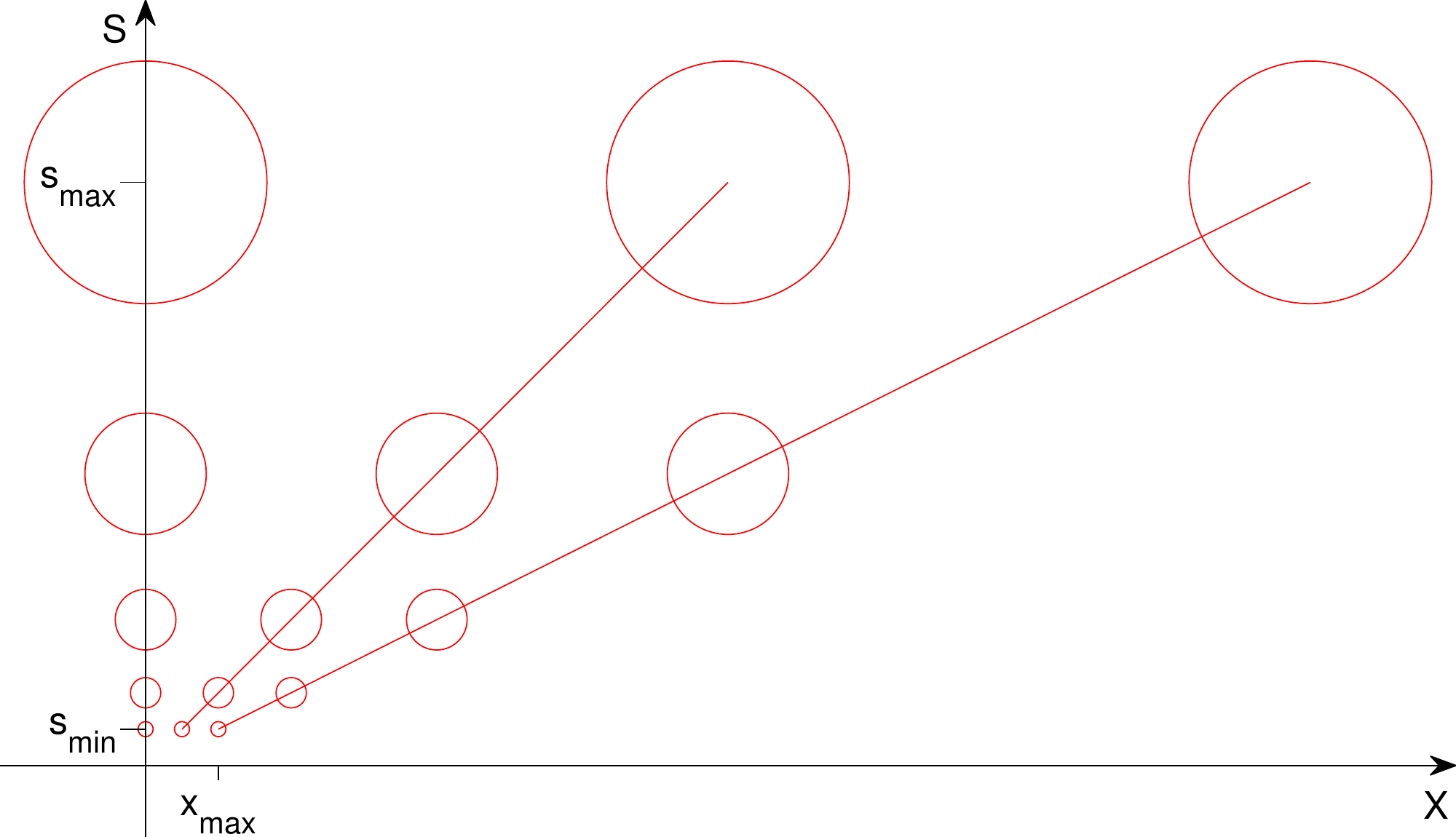}
\caption{\it Under the assumption of Gabor filters and associated sampling
  for each scale at spatial intervals $\Delta x = s$, this graph
  depicts a subset of the
  resulting array of template units. Note that in this graph the
  sampling over $s$ is arbitrary but the sampling over $x$ follows the
sampling theorem. Note that there will be no samples between the $s$ axis and the line with
slope $1$ (when $x$ and $s$ are plotted in the same units). The center
of the circles in the figure gives the $s,x$ coordinates; the circles
are icons symbolizing here the receptive fields.}
\label{Sampling}
\end{figure}

In the following we {\it assume for simplicity} that the template is a
Gabor filter (of one orientation; other templates may have different
orientations). We assume that the Gabor filter and its transforms
under translation and scaling are roughly bandpass and the sampling
interval at one scale over $x$ is $s$, implying half overlap of the
filters in $x$.  This is illustrated in Figure \ref{Sampling}.

The above assumptions imply that for each array of filters of size $s$
the first unit on the right of the central one (at $x=0$) is at $x=s$,
{\it where $x$ and $s$ are measured in the same units}.

For the scale axis we follow the sampling pattern estimated by Marr et
al.\cite{{Marretal1980}} with $5$ ``frequency channels'' having $2s=$ 1'20'', 3.1',
6.2', 11.7', 21'. Filter channels as described above are supported by
sampling by photoreceptors that starts in the center of the fovea at
the Shannon rate, dictated by the diffraction-limited optics with a
cutoff around $60$ cycles/degree, and then decreases as a function of
eccentricity.

% JM: these aren't evenly spaced in log scale? yes and not evenly
% spaced in linear scale

\subsection{Prediction: size of the foveola and slope of the magnification factor}
\label{sec:M}

In this ``truncated pyramid'' model of V1 (we assume that the Gabor
templates correspond to simple cells in V1, called S1 in HMAX) the
slope of the magnification factor M as a function of eccentricity
depends on the size of the foveola -- that is the region at the minimum
scale $s_{min}$. The larger the foveola, the smaller the slope.

We submit that this ``engineering model'' nontrivially fits data about
the size of the fovea, the slope of M and other data about the size of
receptive fields in V1. In particular the size of the foveola, the
size of the largest RFs in AIT and the slope of acuity as a function
of eccentricity depend on each other: fixing one determines the other
(after setting the range of spatial frequency channels, i.e., the
range of RF sizes at $x=0$ in V1).

\subsubsection{Back of the envelope estimates: size and sampling of
  foveola and fovea}

% JM: just a note to summarize several things to take into consideration in order to estimate the slope:
% - if we're going to be using s_min expressed as an RF radius, then we'll need slope expressed in (RF radius / eccentricity), not (RF size / eccentricity)
% - monkey data or human data?
% - we want the minimum slope, not the average slope
% - are we talking about simple cells or complex cells?

\begin{figure}\centering
\includegraphics[width = 0.8\textwidth]{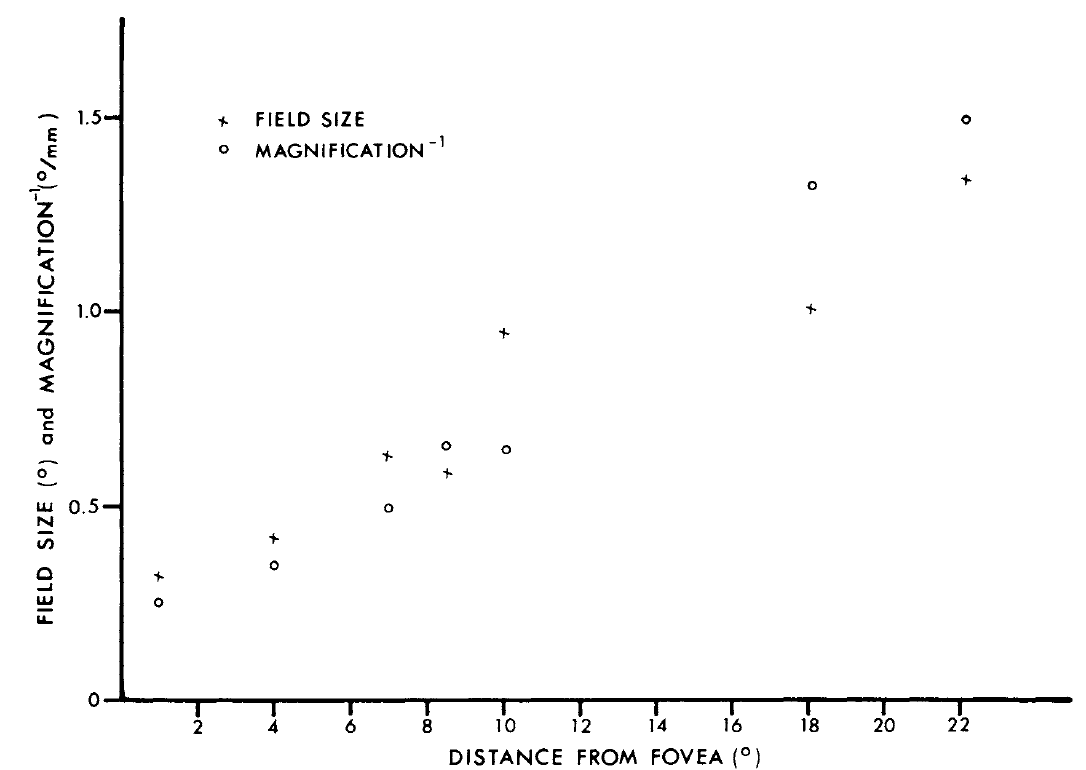}
\caption{\it Data of Hubel and Wiesel for monkey V1 gives a slope for average RF diameter, relative to eccentricity, of $a=0.05$.  From \cite{Hubel1974}.}
\label{hubel74}
\end{figure}

\begin{figure}\centering
\includegraphics[width=0.8\textwidth]{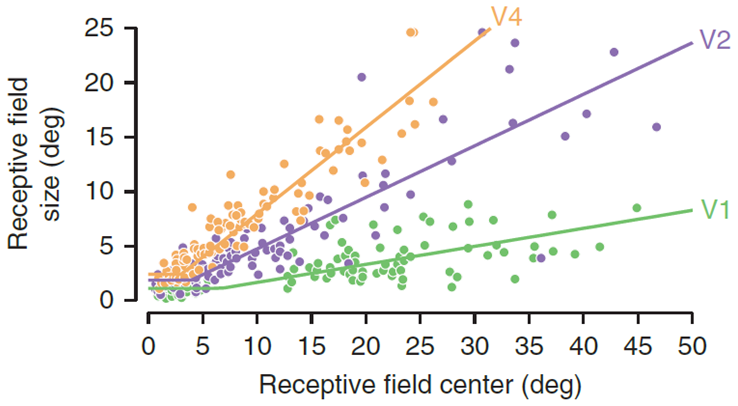}
\caption{{\it V1 data supports a model in which the {\em maximum} scale also depends on eccentricity, as in figure~\ref{VandV} (lower).  Adapted from \cite{Freeman2011} (original monkey data from \cite{Gattass1981,Gattass1988}).  Note that the y-axis here represents scale using RF diameter, not spacing.}}
\label{freeman}
\end{figure}

\begin{figure}\centering
\includegraphics[width = 0.6\textwidth]{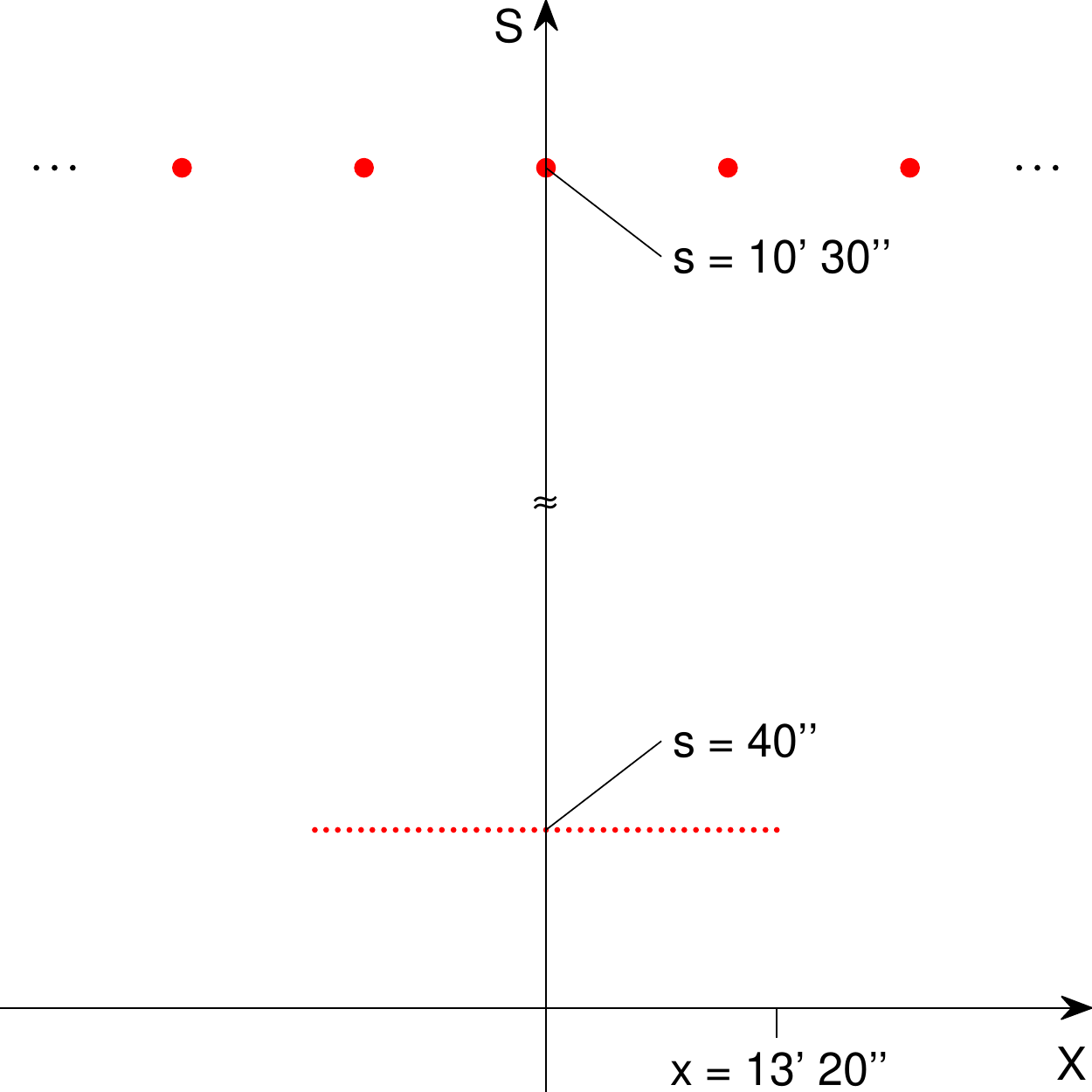}
\caption{\it The figure shows a foveola of $\approx 26'$ and a total of $\approx 40$
  units ($20$ on each side of the center of the fovea). It also shows
  sampling intervals at the coarsest scale in the fovea (assumed to be
  around $2s=21'$\cite{Marretal1980} which would span $\approx \pm
  6^\circ$.
Note that the size of a letter on the ophthalmologist's screen
  for 20/20 vision is around $5'$.}
\label{Sizefovea}
\end{figure}

As a back of the envelope calculation we assume here that $s_{min}
\approx 40"$ (from an estimate of $1'20''$ for the diameter of the
smallest simple cells \cite{{Marretal1980}}, see also
\cite{Mazer2002}). Data of Hubel and Wiesel \cite{Hubel1962,Hubel1974}
(shown in Figure~\ref{hubel74}) and Gattass \cite{Gattass1981,
  Gattass1988} (shown in Figure~\ref{freeman}) yield an estimate of
the slope $a$ for M in V1 (the slope of the line $s_{min}(x) =
ax$). Hubel and Wiesel cite the slope of the average RF diameter in V1
as $a=0.05$; Gattass quotes a slope of $a \approx 0.16$ (in both cases
the combination of simple and complex cells may yield a biased
estimate relative to the ``true'' slope of simple cells alone). Our
model of an inverted truncated pyramid predicts (using an estimate of
$a=0.1$) from these estimates that the radius of the foveola (the
bottom of the truncated pyramid) is $R= 1'20/0.1 \approx 13'$ with a
full extent of $2R \approx 26'$ corresponding to about $40$ cells
separated by $40"$ each. The size of the fovea (the top of the
truncated pyramid) would then have $2R \approx 6^\circ$ with $40$
cells spaced $\approx 21'$ apart; see Figure \ref{Sizefovea}.

{\it These estimates depend on the actual range of receptive field sizes
and could easily be wrong by factors of $2$. } We have used liberally
data from the macaque together with data from human psychophysics.
Our main goal is to provide a logical interpretation of future data and a
ballpark estimate of relevant quantities.

\subsection{Sampling array in V1}
\label{sec:sampling}

\begin{figure}\centering
\includegraphics[width = 0.6\textwidth]{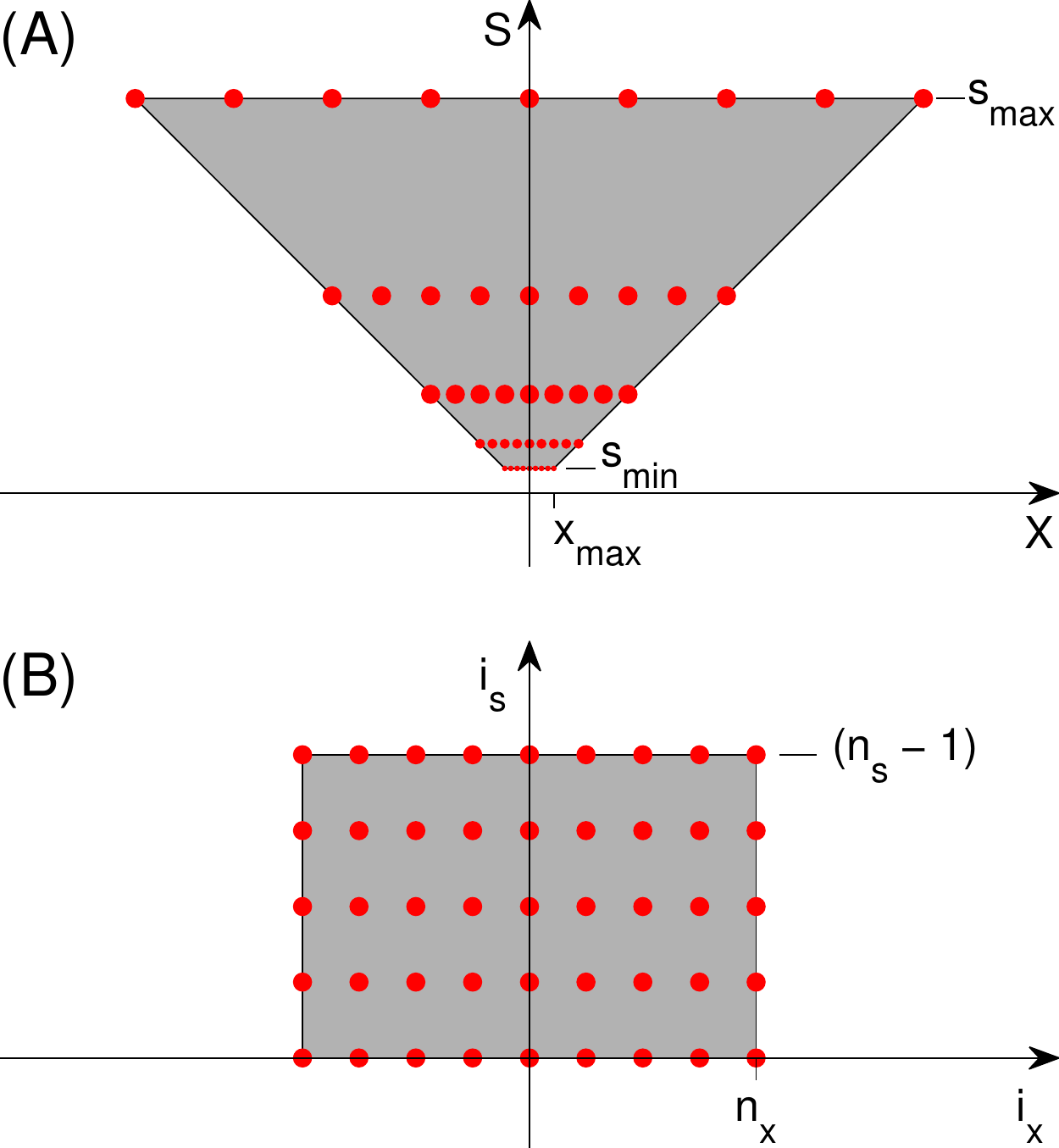}
\caption{\it Since the inverted truncated pyramid (A) has the same number of
sample points at every scale, it maps perfectly onto a square
array (B) when $x$ is replaced by $i_x = x/s$, i.e., the number of samples from the center.
$i_s$ is the scale band number. We call this the ``magic map''.}
\label{SquareArray}
\end{figure}

It is possible to map the nonuniform sampling array into a square
lattice; see Figure \ref{SquareArray}.  (It is actually a cube when
$y$ is included; note that a further mapping of this cube onto a 2D
cortical sheet would be similar to -- but different from! --- the
roughly log-polar mapping commonly assumed to take place between the
retina and V1.)

% JM: 'conformal' applies to a mapping from 2D->2D, this is 2D->3D.

Once the $s,x$ array has been transformed into a square lattice, the
lattice behaves like $x,y$ from the point of view of group
transformations, that is shifts and ``scaling'' commute. An array
spanning $26'$ at the bottom and $6^\circ$ at the top is a kind of
image, but in pixel and scale space.

Each of the scale layers has the same number of units which is
determined by the number of units in the fovea -- that is, the number
of units at the finest resolution.  This remapping shows that S1
corresponds to a lattice of dimensions $x,y,s,\theta$, where the
dimension sizes are different (but roughly the same for $x, y$); the
topology is that of a cylinder with $\theta$ being periodic.

Coordinates $(i_s,i_x)$ within the square lattice are obtained from $(s,x)$ as follows:
$$i_s = log_f(\frac{s}{s_{min}})$$
$$i_x = \frac{x}{s}$$
\noindent where $f$ is the multiplicative factor between adjacent
scale bands.  $i_s$ is the number of the scale band, starting with $0$
for the finest band.  $i_x$ is the number of samples (in position)
from the center, and may be positive or negative.  The lattice will
have size $n_s$ in scale and $2 n_x + 1$ in position.

\begin{figure}\centering
\includegraphics[width = 0.6\textwidth]{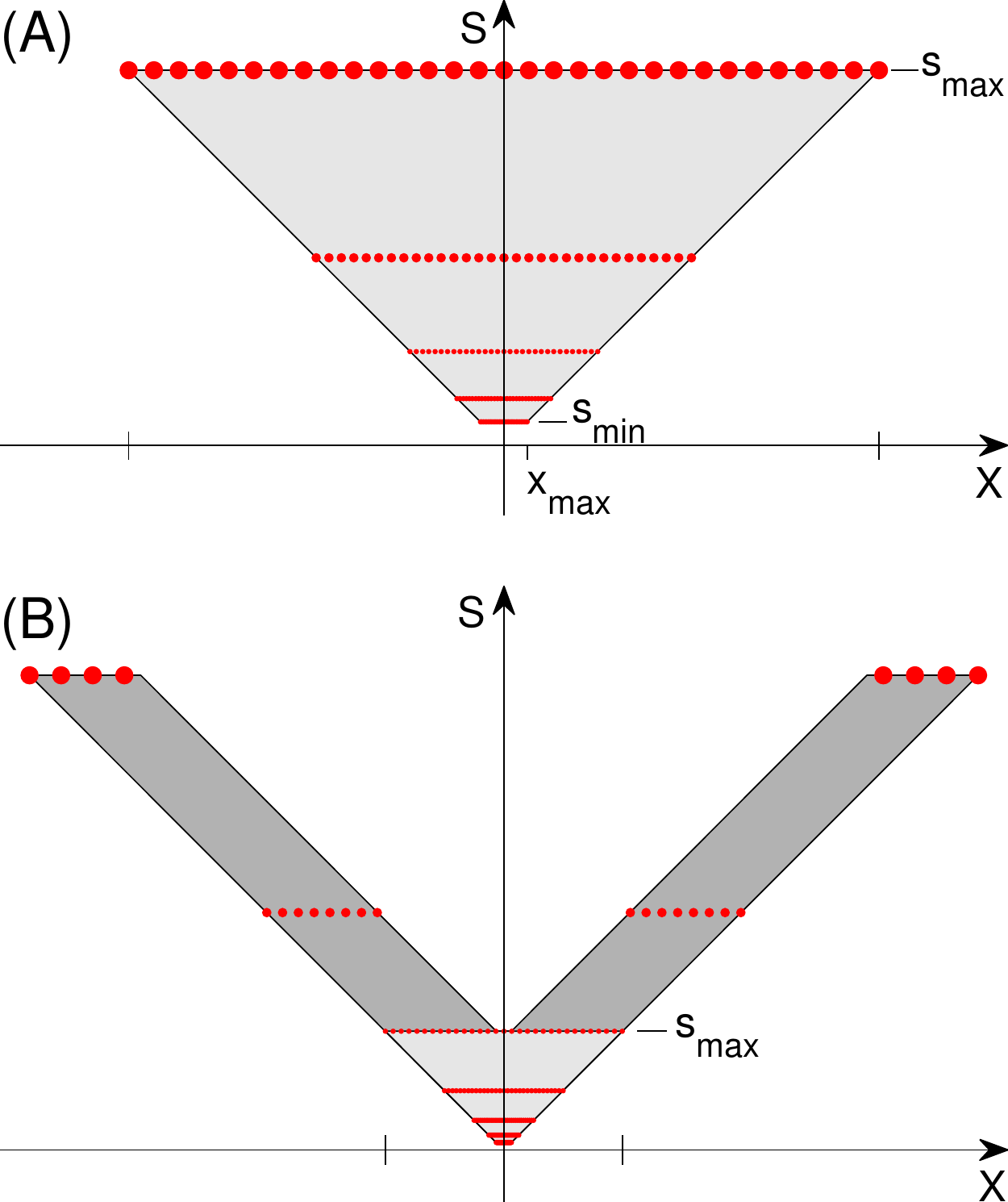}
\caption{\it Regions of scaled and shifted templates.  In the upper plot we have a constant maximum scale ($s_{max}$).
In the lower plot we consider a wider range of eccentricities, and there is a constant {\em difference} between the
largest and smallest scale at each eccentricity.  Note that the central region in the lower plot (lighter shading) is
the same one shown in the upper plot.}
\label{VandV}
\end{figure}

As we have described, the region of Figure \ref{VandV} (upper) is
obtained by setting a minimal scale $s_{min}$, shifting the template
between $-x_0$ and $x_0$ and then scaling all the templates obtained
so far up to $s_{max}$. One of many alternatives is to set a constant
difference between the minimum and the maximum scale at each
eccentricity ($s_{max}-s_{min}=const.$); see Figure \ref{VandV}
(lower). Experimental data suggests this is a more likely possibility
(large eccentricities are also represented in the visual system); see
Figure \ref{freeman}, though several variations are of course equally
possible.

%JM: does the term conformal apply?
%JM: changed above from saying the diagonal boundaries are *required*

%\begin{figure}\centering
%\includegraphics[width = 0.8\textwidth]{s-x_group.jpg}
%\caption{s-x group}
%\label{sxgroup}
%\end{figure}

\subsection{Pooling in C1}
Invariance is not provided directly by the array but by the pooling
over it.  Note that we are limiting ourselves in this section to the invariant recognition of {\it isolated objects}.

% Figure \ref{poolinginvariance} illustrates two extreme
% examples of pooling -- over $x$ for each $s$ or over $s$ for each
% $x$. The psychophysical consequence of such pooling schemes (assuming
% a 1-layer architecture) would be limited generalization. For example,
% in the case of pooling over $s$ for each $x$ (Figure
% \ref{poolinginvariance} plot B), generalization would be from one
% scale to all others but not across positions.  Figure \ref{scaleline}
% illustrates another limited pooling scheme, this time to scaling
% starting from a specific size and position.

% \begin{figure}\centering
% \includegraphics[width = 0.6\textwidth]{figs/typespooling.pdf}
% \caption{\it Two extreme cases of pooling. (A) depicts pooling only over $x$
% for each $s$, (B) pooling over $s$ for each $x$.  Either would yield
% limited invariance and thus poor generalization.}
% \label{poolinginvariance}
% \end{figure}

% \begin{figure}\centering
% \includegraphics[width = 0.6\textwidth]{figs/scaleinvariance.pdf}
% \caption{\it Another pooling scheme yielding limited invariance, in this case,
% to scaling starting from a specific size and position.}
% \label{scaleline}
% \end{figure}

The range of invariance in $x$ is limited for each $s$ by the slope of
the lower bound of the inverted pyramid (see Figure
\ref{SquareArray} plot A).  The prediction is that the range of
invariance $\Delta x$ depends on scale $s$ as
$$\Delta x \approx n_x s,$$

\noindent where $n_x$ is the radius of the inverted pyramid in samples, and is
constant for all scales.  Thus small details (high
frequencies) have a limited invariance range in $x$ whereas larger details
have larger invariance.  $n_x$ is obtained from the slope ($a$) of the cortical
magnification factor (section \ref{sec:M}) as:
$$n_x = 1/a$$

%\begin{figure}\centering
%\includegraphics[width = 0.8\textwidth]{invarianceprediction.jpg}
%\caption{Prediction: range in $x$ of translation invariance depends linearly on scale $s$.}
%\label{translationinvariance}
%\end{figure}

%JM: use variables (i_s, i_x, n_s, n_x) from this point on, when referring to the square lattice?

\section{Part II: Hierarchy and decimating the array}

We follow the magic theory's extension to hierarchies of pooling, dot
products, sampling, pooling. At each dot product (S unit) stage or
pooling (C unit) stage, there can be a downsampling of the array
of units (in $x,y$ and possibly in $s$) that follows from the
low-pass-like effect of the dot-product or pooling operation.

Here we assume a specific strategy of downsampling in space by $2$ at
each stage. This choice is for simplicity but is roughly consistent
with biological data. The results below can be easily changed using
different criteria for downsampling. The following statement follows from
considerations of invariance for image patches that are smaller than
half the pooling region:

{\bf Sampling theorem for pooling} {\it Pooling over $2 x 2$ units
  gives local invariance to scale and translation and permits
  downsampling of the array in $x$ by a factor of $2$.}

\begin{figure}\centering
\includegraphics[width = 1.0\textwidth]{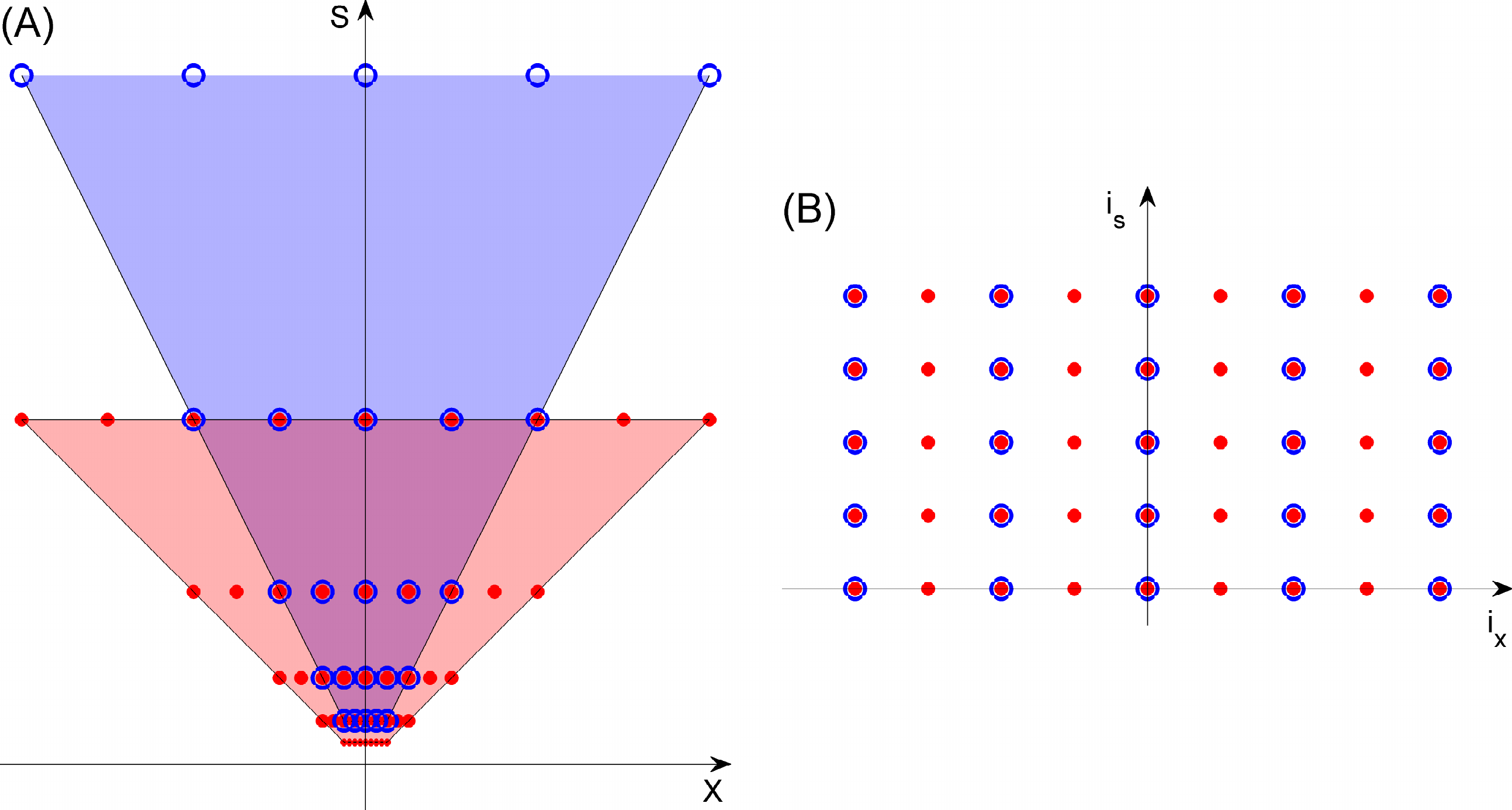}
\caption{\it Pooling over $2x2$ lattices of the $s,x$ array in V1 and
  subsampling reduces the square lattice; if the lattice in $x$ is
  $40$ units, 4 layers (V2, V4, TEO, AIT) of $x$ pooling are sufficient to create
  cells that are 16 times larger than the largest in the fovea in V1
  (probably around $21'$ at the coarsest scale),yielding at the top cells with a RF diameter
  of up to $\approx 5^\circ$. Each area in the fovea would see a
  doubling of size with corresponding doubling of the slopes at
  the border (before remapping to a cube lattice). The index of the
  units at position $x$ and scale $s$ is given by $i^s_x = 2^s i^1_x
  $.  }
\label{2layers}
\end{figure}

\noindent This simple result is discussed elsewhere. We assume here
that each combined S-C stage brings about a downsampling by $2$ of the
$x,y$ array. We call this process ``decimation''; see Figure
\ref{2layers}.  Starting with V1, $4$ stages of decimation (possibly
at V2, V4, TEO and AIT) reduce the number of units at each scale from
$\approx 40$ to $\approx 2$ spanning $\approx 26'$ at the finest scale
and $\approx 6^\circ$ at the coarsest.

Pooling over scale in a similar way may also decimate the array down
to $\approx 1$ scale from V1 to IT. Neglecting orientations, in
$x,y,s$ the $30 \times 30 \times 6$ array of units may be reduced to
just a few units in $x,y$ and $1$ in scale. This picture is consistent
with the invariance found in IT cells
(\cite{serre2005theory}). According to the magic theory, different
types of such units are needed, each for one of several templates at
the top level.

\subsection{Clutter and crowding}

So far our discussion has assumed isolated objects in the image --
which is the usual assumption in most theorems of M-theory. Clutter of
course poses a problem. For instance, pooling over the whole $s,x$
domain of Figure \ref{pyramid} could suffer from the presence of
clutter-induced fragments in any location in the inverted pyramid. A
more natural form of pooling is the (specific) layer-by layer
decimation described in the previous section (see also Figure
\ref{2layers} B). Notice that the corresponding pooling range is
uniform across eccentricities in the coordinates of Figure
\ref{2layers} B. The spatial pooling range on $x$ depends on the area:
for V1 it is the sampling interval between the red dots, for V2 it is
the sampling interval between the blue dots: they should be roughly
equivalent to the radius of the receptive field of the complex cells
in V1 and V2 respectively. We assume now the following reasonable
criterion for pooling to remain unaffected by clutter, that is
interference-free from a flanking object: {\it the target and the
  flanking distractor must be separated at least by the pooling
  distance and thus by a complex cell receptive field}. Under this
assumption, our theory predicts that the critical separation for
avoiding crowding should be

\begin{equation}
\Delta x \approx b x
\end{equation}

\noindent since the RF size increases linearly with eccentricity, with $b$
depending on the cortical area responsible for the recognition signal (see Figure
\ref{2layers} A). Thus the theory ``predicts'' Bouma's law of crowding! \cite{Bouma1970} The
experimental value found by Bouma for crowding of $b \approx 0.4$ suggests that the
area is V2 since this is the slope found by Gattass for the dependence
of V2 RF size on eccentricity. Studies of "metameric" stimuli by Freeman and Simoncelli also implicated V2 in crowding and peripheral vision deficiencies \cite{Freeman2011}.

Simulations by Isik et al. (2011) with an HMAX-type model
confirm the intuition that for the model to be interference-free the
spatial separation between the target object and the flanking object must be $\approx b x$.

\section{Discussion}

\subsection{A magic role for V1}

The most striking result of this paper is that the linear increase of
RFs size with eccentricity, found in all primates, is required by
M-theory to compute a scale and position invariant
representation. Also predicted by the computational theory is the
existence of the foveola and its link with the slope of $M^{-1}$--
though not its (small) size that certainly depends on resource
constraints (such as the size of the optical nerve).

An indirect result with potentially interesting implications is that
the representation at the level of S1 (simple cells in V1) after the
``magic map'' of Figure \ref{SquareArray} corresponds to a discrete
{\it abelian} group for shift and ``scale''. In other words, in this
representation, transformations in $i_s$ and $i_x$ commute. This
property is inherited by higher levels in the network and implies
possibly interesting properties for the covariance function associated
with the image representation at each level and for its
eigenfunctions.

\subsection{A framework for understanding ``everyday'' human recognition}

This state of affairs means that there is a quite limited ``field of
vision'' in a single glimpse. Most of an image of 5 by 5 degrees is
seen at the coarsest resolution only, if fixation is in its center; at
the highest resolution only a fraction of the image (up to 30') can be
recognized, and an even smaller part of it can be recognized in a
position invariant way (the number above are rough estimates).

We introduced earlier the idea of ``IP fragment'' corresponding to the
information captured from a single fixation of an image. Such a
fragment is supported on a domain in the 2D space $x,s$, contained in
the inverted truncated pyramid of Figure \ref{pyramid}.  There are two
interesting remarks:

\begin{enumerate}

\item For normally-sized  images and scenes with fixations well
  inside, the resulting IP-fragment will occupy the full spatial and
  scale   extent of the inverted pyramid. Consider now the fragment
  corresponding to an object to be stored in memory (during learning)
  and recognized (at run time). Suppose the best situation for the
  learning stage (the other cases can be discussed in terms of the
  figure): the object is close so that both coarse and fine scales are
  present in its fragment. At run time then the object can be
  recognized whenever it is closer or farther away. The important
  point is that a look at this and the other possible situations (at
  the learning stage) suggest that the matching should always weight
  more the finest available frequencies (bottom of the pyramid). This
  is the finding of Schynz \cite{Schyns2003}. As implied by his work, top-down effects
  may modulate somewhat these weights (this could be done during
  pooling) depending on the task.

\item Assume that such a fragment is stored in
memory for each fixation of a novel image. Then there is the following
trade-off between learning and run-time recognition:

\begin{itemize}
\item if a new object is learned from a single fixation, recognition
  may require multiple fixations at run time to
  match the memory item (given its limited position invariance).
\item if a new object is learned from multiple fixations, with
  diffrent fragments stored in memory each time, run time recognition
  will need a lower expected number of fixations.

\end{itemize}

Notice that the above argument critically rely on ``large'' and
position-independent scale invariance: scale does not need to be
sampled -- unlike position.

The fragments of an image stored in memory via multiple fixations
could be organized into an egocentric map. Though the map is {\it not
  directly used } for recognition (this is a conjecture that should be
experimentally verified), it is {\it probably needed} to plan
fixations during the learning and especially during the recognition
and verification stage (ant thus indirectly used for recognition in
the spirit of minimal images and related recent work by Ullman and
coworkers). Such a map may be somewhat related to Marr's $2
\frac{1}{2}$ sketch, for which no neural correlate has been found as
yet.

\end{enumerate}

\subsubsection{IP-based recognition: a new paradigm for computer
  vision?}
Learning and recognition relying on a set of IP fragments (see the
tradeoff discussed above and its context) may provide a new powerful
framework for object recognition. Computer vision seems so far to
mostly rely on matching only the coarsest scale available in the
pyramid resulting in performance which is different from human
performance.

\subsection{ Some predictions}

We collect here the key predictions of the theory:

\begin{figure}\centering
\includegraphics[width = 0.4\textwidth]{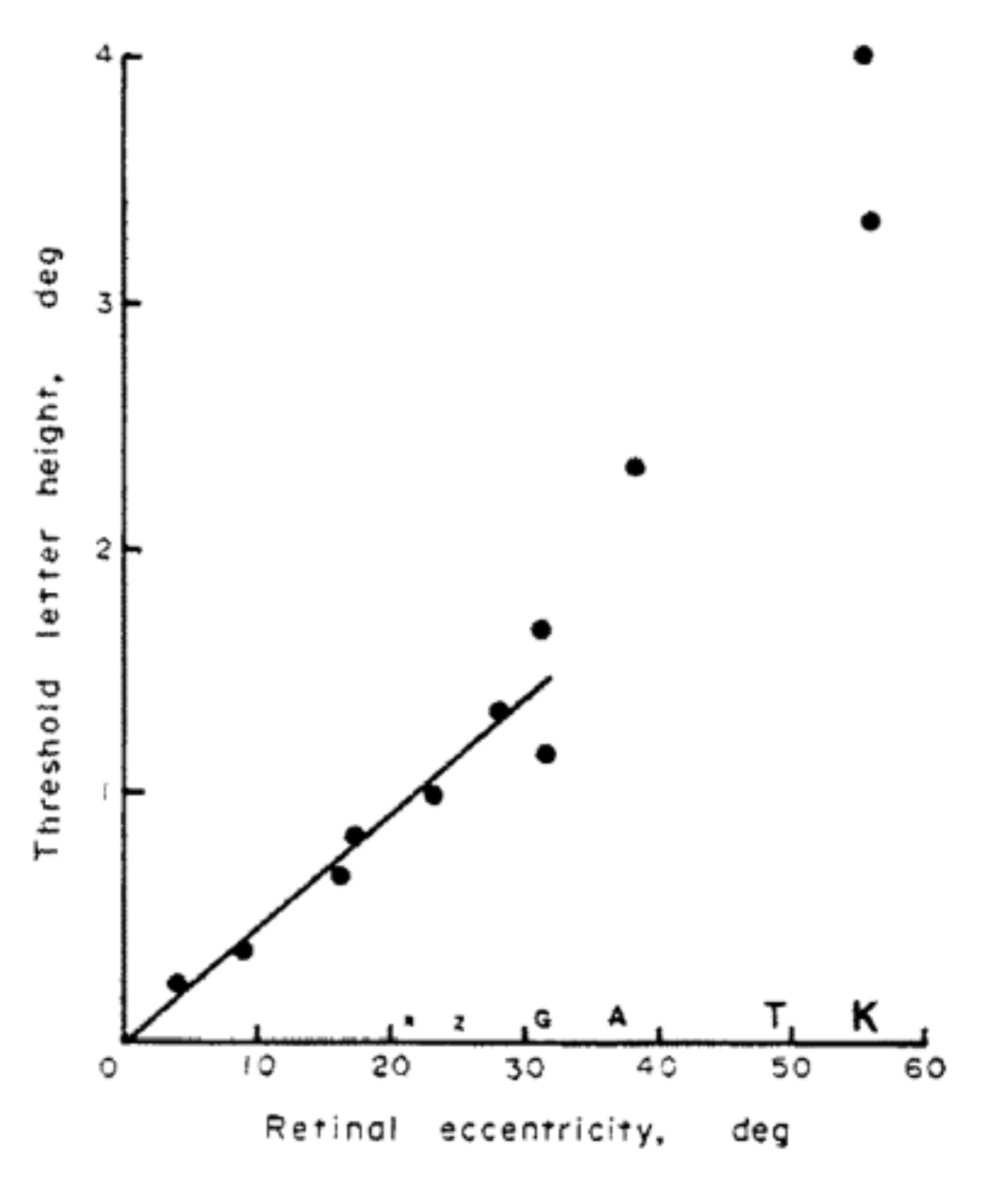}
\caption{{\it Linear size dependency of letter recognition (Anstis, 1974 \cite{Antsis1974}).}}
\label{Anstislinear}
\end{figure}

\begin{figure}\centering
\includegraphics[width = 1.0\textwidth]{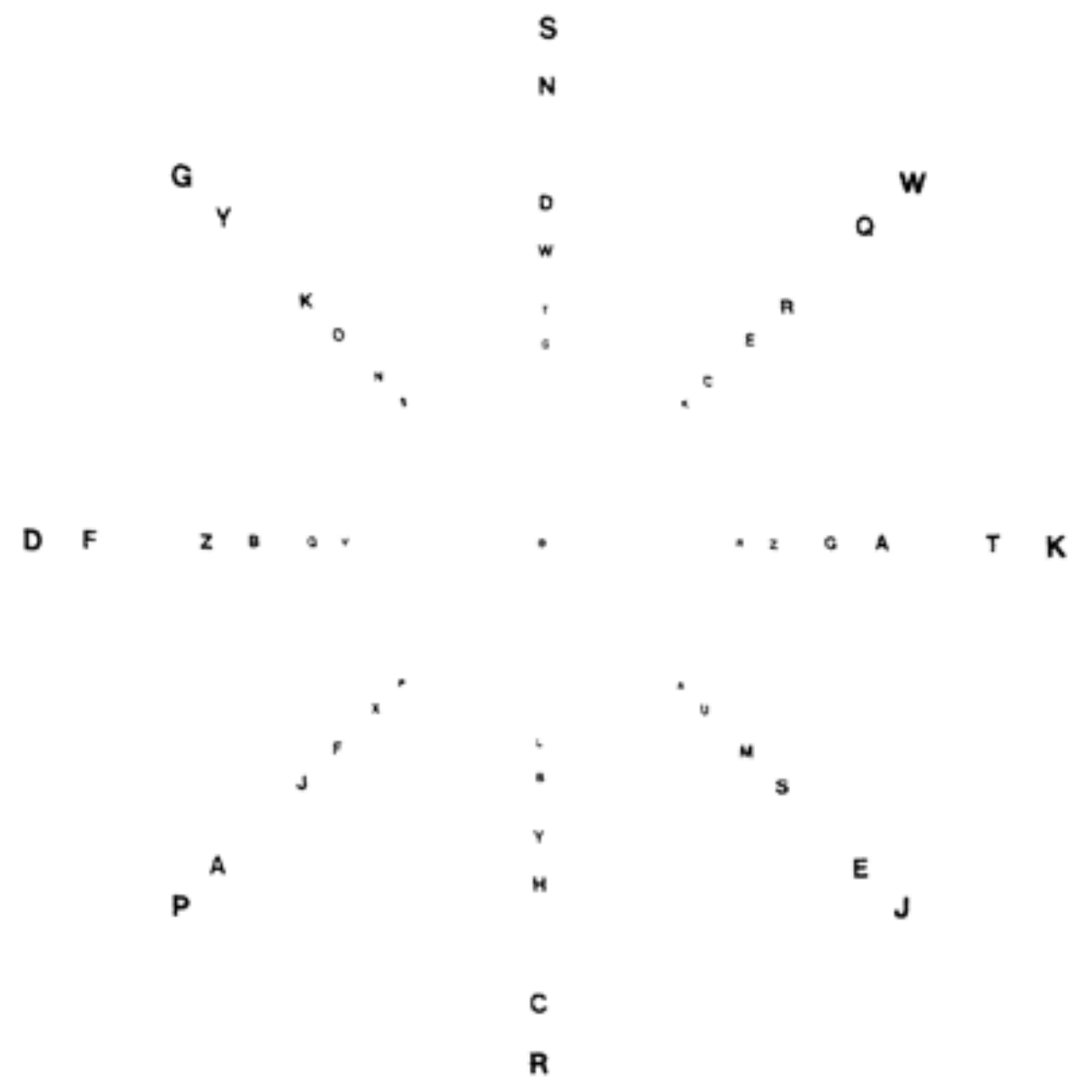}
\caption{{\it When fixating the central point, recognizability of a letter in the chart does not change under scaling (Anstis, 1974).}}
\label{Anstisletters}
\end{figure}

\begin{itemize}

\item  The computational requirement of scale invariance implies linear
  dependency on eccentricity of receptive field sizes; it also
  predicts that the size of the foveola depends on the slope.

\item Image patches that are recognized at some scale and
  eccentricity remain equally recognizable under scaling between
  $s_{min}$ and $s_{max}$. This is a computational prediction of
  Anstis' findings \cite{Anstis1974}. Anstis found (see Figures
  \ref{Anstislinear}, \ref{Anstisletters}) that a letter just
  recognizable at some eccentricity remains equally recognizable under
  scaling: the size of receptive fields in V1 depends on eccentricity
  exactly to match the effect of scaling.

\item Anstis did not measure close to the minimum letter size -- which
  is around $5'$ for 20/20 vision. Our prediction is that if there is
  a range of receptive fields in V1 between $S_{min}$ and $s_{max}$ in
  the fovea then there is a finite range of scaling between $s_{min}$
  and $s_{max}$ under which recognition is maintained (see
  \ref{VandV}). It is obvious that looking at the image from an
  increasing distance will at some point make it unrecognizable; it is
  somewhat less obvious that getting too close will also make it
  unrecognizable (this phenomenon was found in Ullman's minimal
  images; Ullman, personal comm.)

\item Consider the experimental use of images such as letters of
  appropriate sizes that are bandpass filtered (with the filters
  mentioned in the text for human vision). Our predictions --
  extending Anstis' findings -- are that {\it there will be scale
    invariance for all frequencies between $s_{min}$ and $s_{max}$;
    there will be shift invariance that increases linearly with
    spatial wavelength and is at any spatial frequency at least
    between $x_{min}$ and $x_{max}$ (the bottom edge of the truncated
    pyramid). }

\item The theory predicts a flat region of constant maximum resolution
  -- which we suggest may correspond to the so-called {\it
    foveola}. This is different from the empirical equations that have
  been proposed in the past but it is unclear whether the difference
  is experimentally detectable, especially because the biological
  implementation may well be an approximation of the theory.

\item The theory predicts that the slope of the line of minimum $s$
  for each $x$ depends on the size of the foveola and
  vice-versa. Since the slope can be estimated relatively easily from
  a number of existing data, our prediction for the linear size of the
  foveola is around $40$ minutes of arc, corresponding to about $30$
  simple cells of the smallest size (assumed to be $\approx 1'20''$ of
  arc). Notice our definition of the fovea is in terms of the set of
  all scaled versions of the foveola between $s_{min}$ and $s_{max}$
  spanning about $6$ degrees of visual angle.

\item The theory predicts crowding effects due to pooling. It predicts
  Bouma's law and its linear dependency on eccentricity \cite{Bouma1970}.

\item Since Bouma's constant has a value of about $0.4$ (see also
  \cite{Freeman2011}), our theory predicts that an interference-free
  signature from V2 is important for recognition. This is is
  consistent with M-theory independent requirement \cite{MM2013} that
  signals associated with image patches of increasing size are
  accessing memory and classification stages separately. The
  requirement follows from the need of recognizing ``parts and
  whole''. The V2 signal could directly or indirectly (via IT or V4
  and IT) reach memory and classification.

\item The angular size of the fovea remains the same at all stages of
  a hierarchical architecture (V1, V2, V4...), but the number of units
  per unit of visual angle decreases and the slope increases because
  the associated $s$ increases (see Figure \ref{2layers}).

\item Top-down control signals may be able to control the extent of
  pooling, or the pooling stage, used in computing a signature from a
  region of the visual field. This process may correspond to
  attentional suppression of the non-attended region.

\item If we assume that some AIT neurons effectively pool over ``all'' positions
  and scales their invariant receptive field
  (over which consistent ranking of stimuli is maintained) should be
  smaller with higher frequency patterns than with low frequency
  ones. This prediction is consistent with the fact that some
  physiology data suggest that AIT neurons are somewhat less tolerant
  to position changes of small stimuli (Op de Beeck and Vogels 2000 \cite{opdebeek2000},
  also described in \cite{DM2003} ). A comparison across studies suggests
  that position tolerance is roughly proportional to stimulus
  size\cite{DM2003}.
\end{itemize}

\vspace{0.25in}

{\it Acknowledgments}
Cheston Tan suggested the connection with the DiCarlo and Maunsell paper.

\newpage

{\bf Appendix: Notes and Remarks}

\vspace{0.2in}

We list a few experiments which are not critical predictions but may
be worth considering:

\begin{itemize}

\item Create Anstis charts for different tasks (such as letter
  recognition, face recognition, symmetry detection...could they show
  different slopes and foveal sizes?

\item Do crowding parameters depend on task (say letter recognition
  vs face recognition)? They may provide information about the range
  of pooling at cortical area at level $n$ that is responsible for
  specific task.

\item Which experiments could best answer the empirical question of the shape of the upper boundary of the
  magic window?

\item What is the scale and translation invariance of ``minimal
  images''? It may be interesting to appropriately filter images to match one of the visual frequency
  channels.
\item What are the eigenfunctions of the covariance induced by natural
  images at the level of S1, that is after the ``magic map''?

\item Can we measure the size of the foveola by using circular pattern with letters of different sizes?

\item Psychophysics on position invariance did not yet provide a clear
  picture. Our theory suggests that this may be due to experimtal
  designs that have ignored scale and spatial frequency
  variables. Could then sharper results be obtained by repeating
  experiments such as Nazir and O'Regan -- testing position {\it and}
  scale invariance -- with patterns which are bandpass filtered in
  spatial frequency?

\end{itemize}

We list a few remarks:

\begin{itemize}

\item Insead of  bandpass filtering patterns in spatial frequency it
  may be possible to use spatial noise localized in space and
  frequency to block specific frequency channels (this is the dual of
  Schyns approach).

\item {\it In a sense} evolution has given primacy to invariance to
  scale with respect to invariance to translation. A reason for it
  seems obvious for the human eye. Could this be tested considering
  other species with different tradeoffs between moving gaze and
  translating the body?

\item Interestingly, the largest images recognizable at a glance map
  onto about $30 \times 30$ simple cells at the finest resolution,
  spanning $21'$.  (As a remark, the eye of Drosophila contains about
  900 ommatidia.)  They can be scaled up to span $\approx 4^\circ$ at
  the coarsest resolution without changing the associated number of
  bits. There may be a connection with the framework of ``minimal
  images'' of Ullman. In terms of the theory there is a maximum image
  size at each resolution that can be recognized at a glimpse and be
  invariant to scaling, corresponding to the $S$ cell size at that
  level of the hierarchy; the $C$ cell pooling range provides
  invariance over that range. Notice that this state of affairs is
  independent of a multilevel hierarchical architecture; it follows
  from the {\it image hypercube} representation (see Figure
  \ref{SquareArray}) at each level (e.g., S1).

\item Discrete transformations in $i_s$ and $i_x$ commute in the S1 layer
after the ``magic map''  (see Figure \ref{SquareArray}). The property
is maintained at higher layers.

\item The size of the templates at the first layer (S1) determines the size
of the minimal image patch. Of course patterns that are not very
simple (eg bars)  require patches that are at least two or three times the minimal
$1'20''$ (like letters on the ophtalmologist table).

\item The pooling range (at the C1 level) affects the range of  tolerance to
  clutter. The theory ensures invariance and uniqueness if all the
  patches that are pooled are without clutter (they may contain the
  object only).

\item The pooling range at the first layer (C1) is likely to be the
  main determinant of the sampling interval at that level and at the
  S2 level.

\item The set of templates for each C1 must be overcomplete; number of
  templates depends on the number of images to be discriminated $n$
  which depends (in an unknown way) on the size of the image
  patch. One may conjecture that the total number of templates for an
  image patch of size $n$ is less in $k$ layers than in $1$ layer.

\item $M$ is in $mm deg^{-1}$ and $M^{-1}$ is in $mm^{-1} deg$.
  $M^{-1}$ is defined as the receptive field size (in degrees pr mm of cortex).

\end{itemize}

\end{document}